\documentclass[10pt,twocolumn,letterpaper]{article}

\usepackage{iccv}
\usepackage{times}
\usepackage{epsfig}
\usepackage{graphicx}
\usepackage{amsmath}
\usepackage{amssymb}
\usepackage{booktabs}
\usepackage{multirow}
\usepackage{bm}
\usepackage[normalem]{ulem}
\usepackage{xcolor}
\usepackage[para]{footmisc}
\usepackage[ruled,vlined]{algorithm2e}
\usepackage[numbers,sort,compress]{natbib}
\usepackage{url}

\definecolor{citecolor}{RGB}{0, 113, 188}

\newcommand{\system}{FrameStack\xspace}

\makeatletter

\newcommand*\wthelper[2]{%
        \hbox{\dimen@\accentfontxheight#1%
                \accentfontxheight#11.3\dimen@
                $\m@th#1\widetilde{#2}$%
                \accentfontxheight#1\dimen@
        }%
}

\usepackage[pagebackref=true,breaklinks=true,letterpaper=true,colorlinks,bookmarks=false,citecolor=citecolor]{hyperref}

\iccvfinalcopy

\ificcvfinal\fi

\begin{document}

\title{VideoLT: Large-scale Long-tailed Video Recognition}
\newcommand\blfootnote[1]{%
  \begingroup
  \renewcommand\thefootnote{}\footnote{#1}%
  \addtocounter{footnote}{-1}%
  \endgroup
}
    
\author{Xing Zhang$^{1*}$ \quad Zuxuan Wu$^{2,3*}$ \quad Zejia Weng$^2$ \quad Huazhu Fu$^4$ \\ 
    Jingjing Chen$^{2,3}$ \quad Yu-Gang Jiang$^{2,3\dagger}$ \quad Larry Davis$^5$ \\
    $^1$Academy for Engineering and Technology, Fudan University, \\ 
    $^2$Shanghai Key Lab of Intel. Info. Processing, School of Computer Science, Fudan University, \\
    $^3$ Shanghai Collaborative Innovation Center on Intelligent Visual Computing, \\
    $^4$Inception Institute of Artificial Intelligence, 
    $^5$University of Maryland}
    
\maketitle

\begin{abstract}
\blfootnote{$^{*}$ Equal contribution.}
\blfootnote{$^{\dagger}$ Corresponding author.}
Label distributions in real-world are oftentimes long-tailed and imbalanced, resulting in biased models towards dominant labels. While long-tailed recognition has been extensively studied for image classification tasks, limited effort has been made for the video domain. 
In this paper, we introduce \textbf{VideoLT}, a large-scale long-tailed video recognition dataset, as a step toward real-world video recognition. 
VideoLT contains 256,218 untrimmed videos, annotated into 1,004 classes with a long-tailed distribution. Through extensive studies, we demonstrate that state-of-the-art methods used for long-tailed image recognition do not perform well in the video domain due to the additional temporal dimension in videos. 
This motivates us to propose   \system, a simple yet effective method for long-tailed video recognition. In particular, \system performs sampling at the frame-level in order to balance class distributions, and the sampling ratio is dynamically determined using knowledge derived from the network during training. Experimental results demonstrate that \system can improve classification performance without sacrificing the overall accuracy. Code and dataset are available at: \url{https://github.com/17Skye17/VideoLT}.
   
\end{abstract}

\vspace{-4mm}
\section{Introduction}

\begin{figure}
    \centering
    \includegraphics[width=\linewidth]{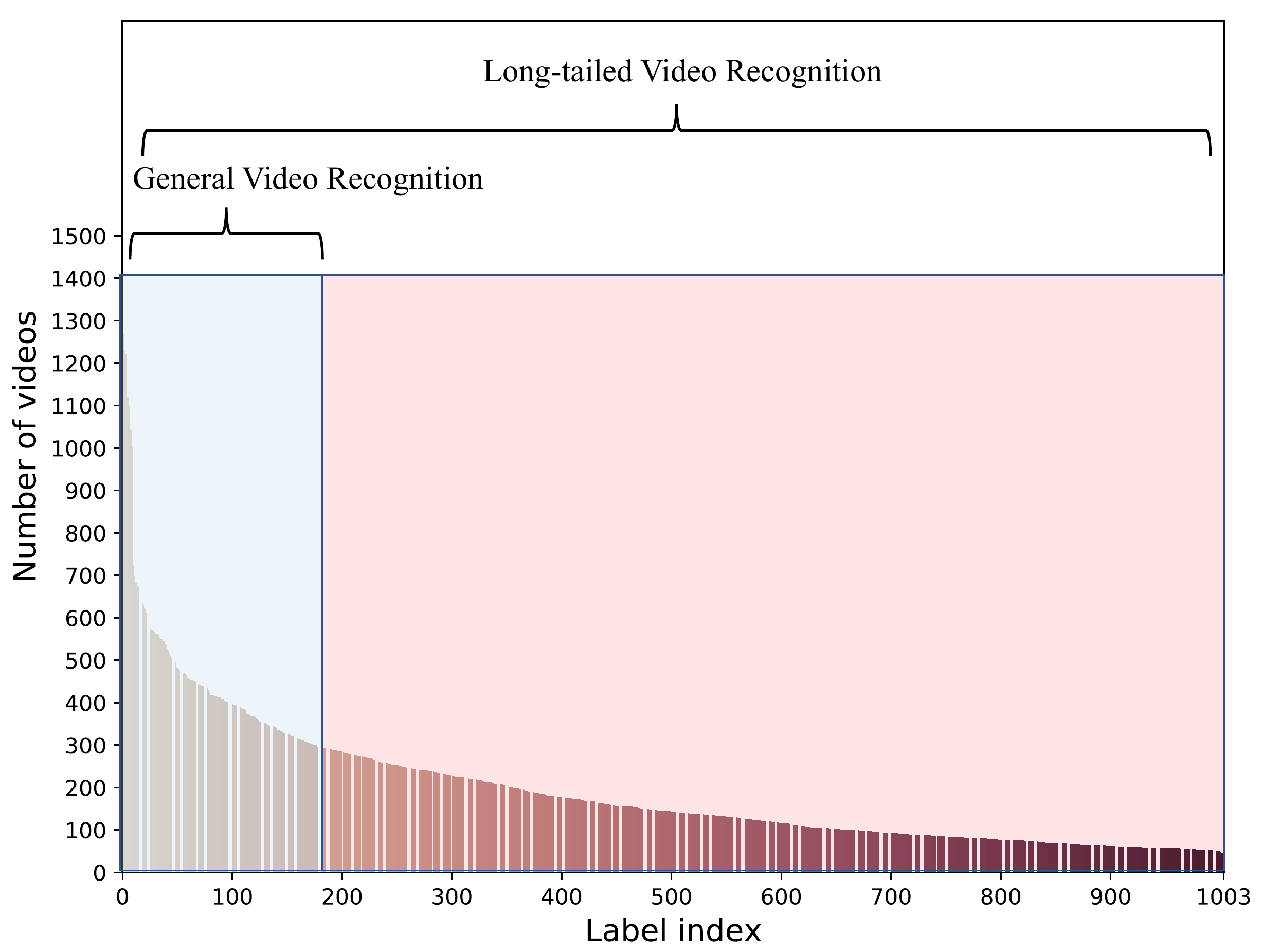}
    \caption{Long-tailed video recognition. General video recognition methods are overfitted on head classes, while long-tailed video recognition focuses on the performance of both head and tail classes, especially on tail classes. (Blue box is the head class region, red box is the region of medium and tail classes.)}
    \label{fig:lt_video}
    \vspace{-4mm}
\end{figure}

Deep neural networks have achieved astounding success in a wide range of computer vision tasks like image classification~\cite{resnet, inception, senet, efficientnet}, object detection~\cite{yolo, ssd, fastrcnn, fasterrcnn}, \etc. Training these networks requires carefully curated datasets like ImageNet and COCO, where object classes are uniformly distributed. However, real-world data often have a long tail of categories with very few training samples, posing significant challenges for network training. This results in biased models that perform exceptionally well on head classes (categories with a large number of training samples) but poorly on tail classes that contain a limited number of samples (see Figure~\ref{fig:lt_video}). 

Recently, there is a growing interest in learning from long-tailed data for image tasks~\cite{decoupling,bbn,tang2020long,eql,yang2020rethinking, dbloss,ldam,cbloss,openLT}. Two popular directions to balance class distributions are re-sampling and re-weighting. Re-sampling~\cite{decoupling,bbn,chawla2002smote,han2005borderline,drumnond2003class} methods up-sample tail classes and down-sample head classes to acquire a balanced data distribution from the original data. On the other hand,  re-weighting methods~\cite{focalloss, rangeloss, ldam, cbloss, eql, dbloss} focus on designing weights to balance the loss functions of head and tail classes. While extensive studies have been done for long-tailed image classification tasks, limited effort has been made for video classification.

While it is appealing to directly generalize these methods from images to videos, it is also challenging since for classification tasks, videos are usually weakly labeled---only a single label is provided for a video sequence and only a small number of frames correspond to that label. This makes it difficult to apply off-the-shelf re-weighting and re-sampling techniques since not all snippets~\footnote{We use ``snippet'' to denote a stack of frames sampled from a video clip, which are typically used as inputs for video networks.} contain informative clues---some snippets directly relate to the target class while others might consist of background frames. As a result, using a fixed weight/sampling strategy for all snippets to balance label distributions is problematic.
For long-tailed video recognition, we argue that balancing the distribution between head and tail classes should be performed at the frame-level rather than at the video (sample)-level---more frames in videos from tail classes should be sampled for training and vice versa. More importantly, frame sampling should be dynamic based on the confidence of neural networks for different categories during the training course. This helps preventing overfitting for head classes and underfitting for tail classes.

To this end, we introduce FrameStack, a simple yet effective approach for long-tailed video classification. \system operates on video features and can be plugged into state-of-the-art video recognition models  with  minimal surgery. More specifically,  given a top-notch classification model which preserves the time dimension of input snippets, we first compute a sequence of features as inputs of FrameStack, \ie, for a input video with $T$ frames, we obtain $T$ feature representations. To mitigate the long tail problem, we define a temporal sampling ratio to select different number of frames from each video conditioned on the recognition performance of the model for target classes. If the network is able to offer decent performance for the category to be classified, we then use fewer frames for videos in this class. On the contrary, we select more frames from a video if the network is uncertain about its class of interest. We instantiate the ratio using running average precision (AP) of each category computed on training data. The intuition is that AP is a dataset-wise metric, providing valuable information about the performance of the model on each category and it is dynamic during training as a direct indicator of progress achieved so far.
Consequently, we can adaptively under-sample classes with high AP to prevent over-fitting and up-sample those with low AP. 

However, this results in samples with different time dimensions and such variable-length inputs are not parallel-friendly for current training pipelines. Motivated by recent data-augmentation techniques which blend two samples~\cite{cutmix,manifoldmixup} as virtual inputs, \system performs temporal sampling on a pair of input videos and then concatenates re-sampled frame features to form a new feature representation, which has the same temporal dimension as its inputs. The resulting features can then be readily used for final recognition. We also adjust the the corresponding labels conditioned on the temporal sampling ratio. 

Moreover, we also collect a large-scale long tailed video recognition dataset, VideoLT, which consists of 256,218 videos with an average duration of 192 seconds. These videos are manually labeled into 1,004 classes to cover a wide range of daily activities. Our VideoLT have 47 head classes ($\#\text{videos} > 500$), 617 medium  classes ($100 < \#\text{videos} <= 500$) and 340 tail classes ($\#\text{videos} <= 100$), which naturally has a long tail of categories. 

Our contributions are summarized as follows:
\begin{itemize}
    \item  We collect a new large-scale long-tailed video recognition dataset, VideoLT, which contains 256,218 videos that are manually annotated into 1,004 classes. \textit{To the best of our knowledge, this is the first ``untrimmed'' video recognition dataset which contains more than 1,000 manually defined  classes.}
    \item We propose \system, a simple yet effective method for long-tailed video recognition. \system uses a temporal sampling ratio derived from knowledge learned by networks to dynamically determine how many frames should be sampled. 
    \item We conduct extensive experiments using popular long-tailed methods that are designed for image classification tasks, including re-weighting, re-sampling and data augmentation. We demonstrate that the existing long-tailed image methods are not suitable for long-tailed video recognition. By contrast, our \system combines a pair of videos for classification, and achieves better performance compared to alternative methods. The dataset, code and results can be found at: \url{https://github.com/17Skye17/VideoLT}.
    
\end{itemize}

\begin{figure*}
\begin{minipage}[t]{0.48\textwidth}
    \centering
    \includegraphics[width=9cm]{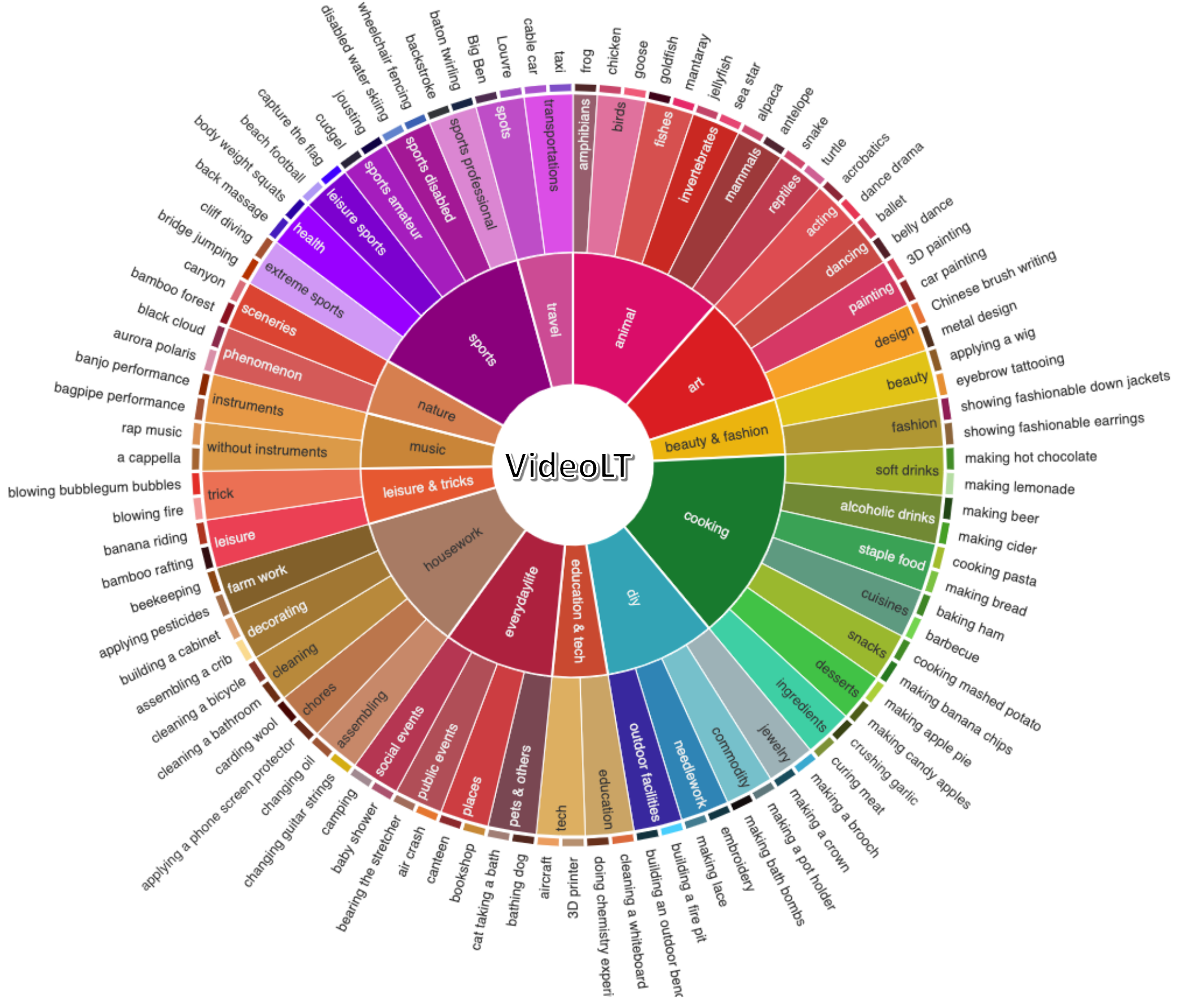}
    \caption{The taxonomy structure of VideoLT. There are 13 top-level entities and 48 sub-level entities, the children of sub-level entities are sampled. Full taxonomy structure can be found in Supplementary materials.}
    \label{fig:taxonomy}
    \vspace{-3mm}
\end{minipage}
\hspace{4mm}
\begin{minipage}[t]{0.48\textwidth}
    \centering
    \includegraphics[width=8cm]{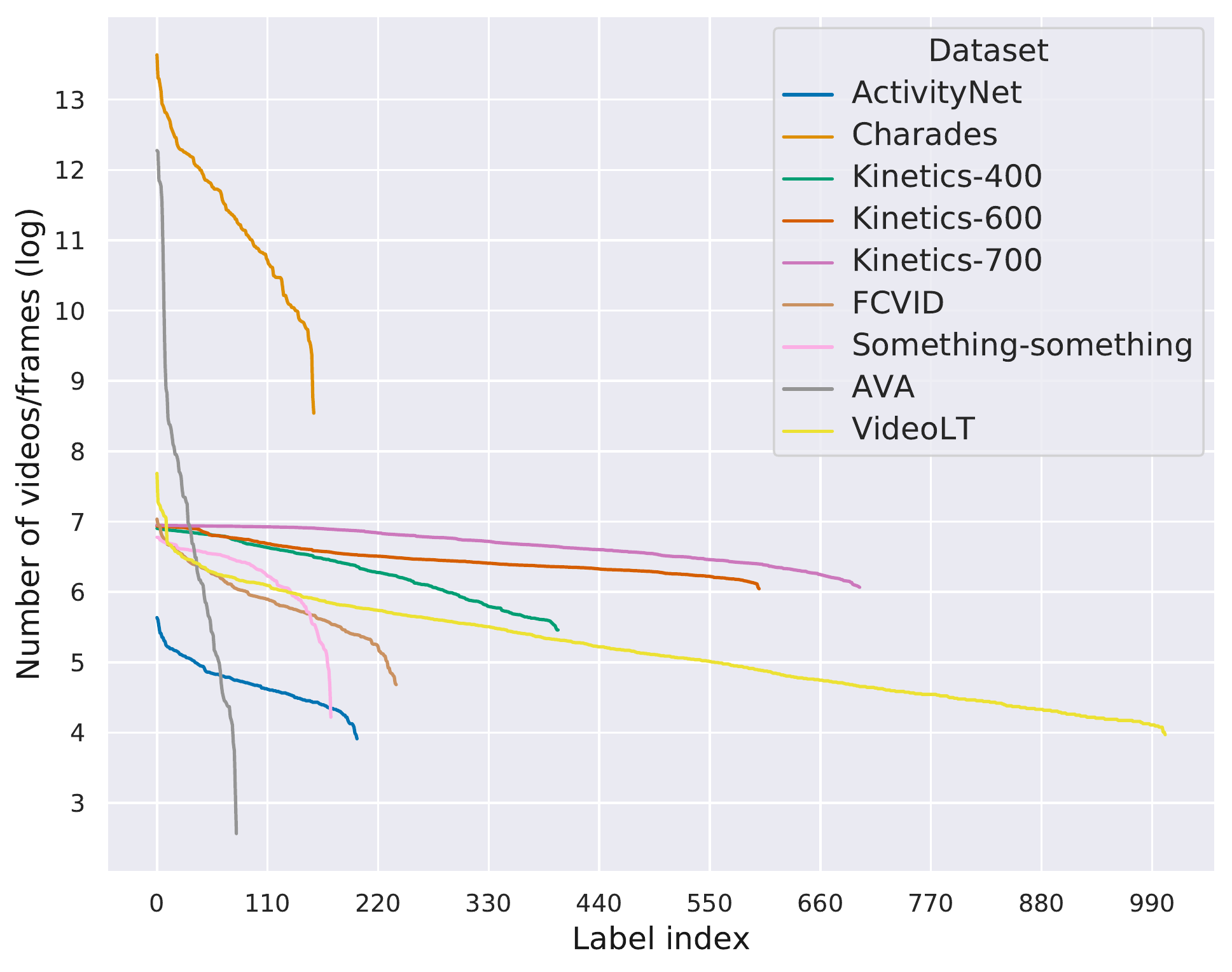}

    \caption{Class frequency distribution of existing video datasets and VideoLT. VideoLT has superior linearity in logarithmic coordinate system, which means the class frequency distribution of VideoLT is close to long-tailed distribution.}
    \label{fig:vdata}
    \vspace{-3mm}
\end{minipage}
\end{figure*}

\section{Related Work}
\subsection{Long-Tailed Image Recognition}
Long-tailed image recognition has been extensively studied, and there are two popular strands of methods: re-weighting and re-sampling.

\vspace{0.05in}
\noindent
\textbf{Re-Weighting}  A straightforward idea of re-weighting is to use the inverse class frequency to weight loss functions in order to re-balance the contributions of each class to the final loss. However, the inverse class frequency usually results in poor performance on real-world data~\citep{cbloss}. To mitigate this issue,  Cui \etal \cite{cbloss} use carefully selected samples for each class to re-weight the loss. Cao \etal \cite{ldam} propose a theoretically-principled label-distribution-aware margin loss and a new training schedule DRW that defers re-weighting during training. In contrast to these methods, EQL~\cite{eql} demonstrates that tail classes receive more discouraging gradients during training, and ignoring these gradients will prevent the model from being influenced by those gradients.

For videos, re-weighting the loss is sub-optimal as snippets that are used for training contain different amount of informative clues related to the class of interest---assigning a large weight to a  background snippet from tail classes will likely bring noise for training.

\vspace{0.05in}
\noindent
\textbf{Re-Sampling}. There are two popular types of re-sampling: over-sampling and under-sampling. Over-sampling~\cite{chawla2002smote,han2005borderline} typically repeats samples from tail classes and under-sampling~\cite{drumnond2003class} abandons samples from head classes. Recently, class frequency is used for class-balanced sampling~\cite{ca-sampling,mahajan2018exploring, bbn, decoupling}. 
BBN~\cite{bbn} points out that training a model in an end-to-end manner on long-tailed data can improve the discriminate power of classifiers but damages representation learning of networks. Kang \etal~\cite{decoupling} show that it is possible to achieve strong long-tailed recognition performance by only training the classifier.
Motivated by these observations~\cite{decoupling,bbn}, we decouple feature representation and classification for long-tailed video recognition. But unlike these standard re-sampling methods, we re-sample videos by concatenating frames from different video clips.

\vspace{0.05in}
\noindent
\textbf{Mixup}. Mixup~\cite{zhang2017mixup} is a popular data augmentation method that linearly interpolates two samples at the pixel level and their targets as well. There are several recent methods improving mixup from different perspectives. For example Manifold Mixup~\cite{manifoldmixup} extends mixup from the input space to the feature space. CutMix~\cite{cutmix} cuts out a salient image region and pastes it to another image. PuzzleMix~\cite{puzzlemix} uses salient signals without removing the local properties of inputs. ReMix~\cite{remix} designs disentangled mixing factors to handle imbalanced distributions and improve the performance on minority classes. Recently, several studies show that mixup is also powerful when dealing with the long tail problem~\cite{remix, bbn}, because it brings higher robustness and smoother decision boundaries to models and it can reduce overfitting to head classes. Our approach is similar to mixup in that we also combine two videos and mix their labels. However, \system operates on frame features along the temporal dimension, but more importantly the mixing ratio in \system is dynamic based on knowledge from the network model.

\subsection{General \& Long-tailed Video Recognition}
Extensive studies have been made on video recognition with deep neural networks~\cite{kinetics, qiu2017learning, feichtenhofer2019slowfast, feichtenhofer2020x3d,wang2018non,lin2019tsm,hussein2019timeception} or training methods~\cite{wu2020multigrid} for video recognition.
These approaches focus on learning better features for temporal modeling, by developing plugin modules \cite{wang2018non,lin2019tsm} or carefully designing end-to-end network structures \cite{feichtenhofer2019slowfast,feichtenhofer2020x3d}. State-of-the-art video recognition models mainly experiment with general video recognition datasets to demonstrate their capacity in modeling long-term temporal relationships \cite{wang2018non,hussein2019timeception} or capturing short-term motion dynamics \cite{kinetics, qiu2017learning,feichtenhofer2020x3d,lin2019tsm}. However, limited effort has been made for long-tailed video recognition due to the lack of proper benchmarks. Zhu and Yang \cite{zhu2020inflated} propose Inflated Episodic Memory to address long-tailed visual recognition in both image and videos. 
However, the use of memory banks in \cite{zhu2020inflated} is resource-demanding. \system is computationally more efficient using in-vitro knowledge from the network for resampling, and thus is more efficient without the need to attend to memory slots. Moreover, \system is a plug-in data augmentation strategy that can be easily applied to model training.

\section{VideoLT Dataset}
We now introduce VideoLT in detail, which is a large scale benchmark designed for long-tailed video recognition.

In contrast to existing video datasets that focus on action or activities~\cite{ucf101, hmdb, sports1m, activitynet, kinetics}, VideoLT is designed to be general and to cover a wide range of daily activities. We manually define a hierarchy with 13 top-level categories including: \textit{Animal}, \textit{Art}, \textit{Beauty and Fashion}, \textit{Cooking}, \textit{DIY}, \textit{Education and Tech}, \textit{Everyday life}, \textit{Housework}, \textit{Leisure and Tricks}, \textit{Music}, \textit{Nature}, \textit{Sports} and \textit{Travel}. See Figure~\ref{fig:taxonomy} for details.
For each top-level class, we used ConceptNet to find sub-level categories. Finally, we selected $1,004$ classes for annotation. To obtain a more diverse video dataset, we not only use the defined categories in taxonomy system, but also expand tags with the same semantics. Then we use these tags to search and crawl videos from YouTube. For each category, duplicate videos and some very long videos are removed, and the number of videos for all categories is larger than 80.  In order to ensure annotation quality, each video is labelled by three annotators and a majority voting is used to determine the final labels. See supplemental materials for more details.

VideoLT is split into a training set, a validation set and  a test set using  70\%, 10\% and 20\% of videos, respectively. To better evaluate approaches for long-tailed recognition, we define 47 head classes ($\#videos > 500$), 617 medium  classes ($100 < \#videos <= 500$) and 340 tail classes ($\#videos <= 100$). See Supple. for details.

\vspace{0.05in}
\noindent
\textbf{Comparisons with existing video datasets}
We visualize in Fig.~\ref{fig:vdata} the class frequency distribution of the training and validation set from ActivityNet v1.3~\cite{activitynet}, Charades~\cite{charades}, Kinetics-400~\cite{kinetics}, Kinetics-600~\cite{kinetics-600}, Kinetics-700~\cite{kinetics-700}, FCVID~\cite{jiang2017exploiting}, Something-something v1~\cite{sth-sth}, AVA~\cite{ava} and VideoLT. VideoLT has superior linearity in logarithmic coordinate system, which means the class frequency distribution of VideoLT is close to a long-tailed distribution.

It is worth noting that YouTube-8M is a large scale dataset with 3,862 classes and 6.8 million videos~\cite{abu2016youtube}. With so many categories, the dataset naturally has a long tail distribution of classes as ours. However, the classes in YouTube-8M are inferred by algorithms automatically rather than manually defined. Each video class has at least 200 samples for training, which is two times higher than ours. In addition, it does not provide head, medium, tail classes for better evaluations of long-tailed approaches.

\section{\system}
We now introduce \system, a simple yet effective approach for long-tailed video recognition. In image recognition tasks, input samples always correspond to their corresponding labels. However, for video recognition, snippets, which might not contain informative clues due to the weakly-labeled nature of video data, could also be sampled from video sequences for training. Popular techniques with a fixed re-sampling/re-weighting strategy for long-tailed image recognition are thus not applicable, since they will amplify noise in background snippets when calculating losses.

To mitigate imbalanced class distributions for video tasks, \system re-samples training data at the frame level and adopts a dynamic sampling strategy based on knowledge learned by the network itself. The rationale behind \system is to dynamically sample more frames from videos in tail classes and use fewer frames for those from head classes.  
Instead of directly sampling raw RGB frames to balance label distributions, we operate in the feature space by using state-of-the-art models that are able to preserve the temporal dimension in videos~\cite{feichtenhofer2019slowfast,lin2019tsm}~\footnote{Most top-notch recognition models do not perform temporal downsampling till the end of networks.}.  This allows \system to be readily used as a plugin module for popular models to address the long-tail problem in video datasets without retraining the entire network. 

More formally, we represent a video sequence with $L$ frames as $\mathcal{V} = \{{\bm f}_1, {\bm f}_2, \ldots, {\bm f}_L\}$, and its labels as $\bm y$.
We then use a top-notch model (more details will be described in the experiment section) to compute features for $\mathcal{V}$, and the resulting representations are denoted as ${\bm V} =  \{{\bm v}_1,{\bm v}_2,\ldots,{\bm v}_L\}$. To determine how many frames should be selected from ${\bm V}$ for training a classifier,  we compute a running Average Pecision (rAP) during training to evaluate the performance of the network for each category on the entire dataset. For each mini-batch of training samples, we record their predictions and the groundtruth. After an epoch, we calculate the ap for each class on the training set. We refer to this metric as running AP since the parameters of the model are changing every mini-batch. While it is not accurate as standard average precision, it provides relative measurement about how the model performs for different classes.  If the model is very confident for certain categories as suggested by rAP, we simply use fewer frames and vice versa.

However, this creates variable-length inputs for different samples in a batch, which is not parallel-friendly for current GPU architectures. In addition, it is difficult to directly translate rAP to the number of samples to be used. To address this issue, \system operates on a pair of video samples, $({\bm V}_i, {\bm y}_i), ({\bm V}_j, {\bm y}_j)$, which are randomly selected in a batch. Based on their ground-truth labels, we can obtain the corresponding rAPs, $rAP_i$ and $rAP_j$ for classes ${\bm y}_i$ and ${\bm y}_j$, respectively. We then define a temporal sampling ratio as:
\begin{equation}
    \beta=\frac{rAP_i}{rAP_i+rAP_j},
\end{equation}
where $\beta$ indicates the relative performance of classes ${\bm y}_i$ and ${\bm y}_j$ by the network so far. Then the number of frames that are sampled from ${\bm V}_i$ and ${\bm V}_j$ are $L_i$ and $L_j$, respectively:

\begin{equation}
    \begin{array}{l}L_i= \lfloor(1-\beta)\times L\rfloor \\  L_j=\lfloor \beta\times L \rfloor \\\end{array}.
\end{equation}
We then produce two new snippets $\widehat{ {\bm V}_i}$ and $\widehat{ {\bm V}_j}$ with length $L_i$ and $L_j$ derived from $ {\bm V}_i$ and $ {\bm V}_j$ respectively through uniform sampling. By concatenating $\widehat{ {\bm V}_i}$ and $\widehat{ {\bm V}_j}$, we  obtain a new sample $\widetilde{\bm V}$ whose length is $L$:
\begin{equation}
    \widetilde {\bm V}=\texttt{Concat}([\widehat{ {\bm V}_i} \,\,;\,\, \widehat{ {\bm V}_j}]). 
\end{equation}

Now $\widetilde{{\bm V}}$ becomes a multi-label snippet containing categories of ${\bm y}_i$
and ${\bm y}_j$. We associate $\widetilde{{\bm V}}$ with a multi-label vector scaled by $\beta$:
\begin{align}
\tilde{\bm y} = (1-\beta)\times {\bm y}_i + \beta \times {\bm y}_j,   
\end{align}
Then, $\widetilde {\bm V}$ and $\tilde{\bm y}$ can then be used by temporal aggregation modules for classification. Note that at the beginning of the training process, recognition accuracies are pretty low for all categories, and thus $\beta$ is not accurate. To remedy this, when 
$(rAP_i+rAP_j) < 1e-5$, we set $\beta$ to 0.5 to sample a half of frames from the two videos. Algorithm~\ref{alg:alg} summarizes the overall training process.

It is worth pointing out that \system shares similar spirit as mixup~\cite{zhang2017mixup}, which interpolates two samples linearly as data augmentations to regularize network training. Here, instead of mixing frames, we concatenate sampled video clips with different time steps to address the long-tailed video recognition problem.  As will be shown in the experiments, \system outperforms mixup by clear margins in the context of video classification. FrameStack can be regarded as a class-level re-balancing strategy which is based on average precision of each class, we also use focal loss~\cite{focalloss} which adjusts binary cross-entropy based on sample predictions.

\SetKwInput{KwInput}{Input}                
\SetKwInput{KwOutput}{Output}              

\begin{algorithm}
\SetAlgoLined
\KwResult{Updated $rAP$ list. Updated model $f_\theta$}
\KwInput{Dataset $D=\{({\bm V}_i,{\bm y}_i))\}_{i=1}^n$. Model $f_\theta$}
Initialize $rAP=0$, $\varepsilon=1e-5$

\tcc{$M$: videos in a mini-batch}
\For{$e\in Max Epoch$}{
    \For{$({\bm V},{\bm y})\in $ $M$}  
    { 
    $({\bm V}_i,{\bm y}_i))$, $({\bm V}_j,{\bm y}_j))\leftarrow $ Sampler($D$, $M$)

    \uIf{$(rAP_i+rAP_j) < \varepsilon$}{
            $\beta=0.5$ 
        }
    \uElse{
        $\beta=\frac{rAP_i}{rAP_i+rAP_j}$ 
        }
    $L_i \leftarrow \lfloor (1-\beta)\times L \rfloor$ 
    
    $L_j \leftarrow \lfloor \beta\times L \rfloor$ 
    
    \tcc{$\widehat{ {\bm V}_i}$, $\widehat{ {\bm V}_j}$: Uniformly sample $L_i$, $L_j$ frames from ${\bm V}_i$, ${\bm V}_j$}    
    
    $\widehat{ {\bm V}_i} \leftarrow Uniform({\bm V}_i\lbrack L_i\rbrack)$    
    
    $\widehat{ {\bm V}_j} \leftarrow Uniform({\bm V}_j\lbrack L_j\rbrack)$       
    
    $\widetilde{\bm V} \leftarrow Concat\lbrack\widehat{ {\bm V}_i},\widehat{ {\bm V}_j}\rbrack$  
    
    $\tilde{\bm y} \leftarrow (1-\beta)\times {\bm y}_i + \beta \times {\bm y}_j$ 
    }
    
    $\mathcal{L}(f_\theta) \leftarrow {\textstyle\frac1M}{\textstyle\sum_{(\widetilde{\bm V},\tilde{\bm y}) \in M}}\mathcal{L}((\widetilde{\bm V},\tilde{\bm y});f_\theta$) 
    
    $f_\theta \leftarrow f_\theta-\delta\nabla_\theta \mathcal{L}(f_\theta)$ 

    \tcc{$rAP$: A list of running average precision for each class} 
    
    $rAP \leftarrow $ APCalculator 
    
    \Return {$rAP$}
}
\caption{Pseudo code of~\system.}
\label{alg:alg}
\end{algorithm}

\begin{table*}[t]
\renewcommand\arraystretch{1.1} 
\begin{center}
    
\renewcommand\tabcolsep{1.2pt}

\begin{tabular}{c|cccccc|cccccc}
\toprule
\multicolumn{1}{l|}{} & \multicolumn{6}{c|}{ResNet-50}                                                                                                                                                                                                                            & \multicolumn{6}{c}{ResNet-101}                                                                                                                                                                                \\\midrule
 LT-Methods                & Overall & \begin{tabular}[c]{@{}c@{}}{[}500,$+\infty$)\\Head \end{tabular} & \begin{tabular}[c]{@{}c@{}} {[}100,500)\\Medium\end{tabular} & \begin{tabular}[c]{@{}c@{}}{[}0,100)\\Tail \end{tabular}              & Acc@1                & Acc@5               & Overall & \begin{tabular}[c]{@{}c@{}}{[}500,$+\infty$)\\Head \end{tabular} & \begin{tabular}[c]{@{}c@{}}{[}100,500)\\Medium \end{tabular} & \begin{tabular}[c]{@{}c@{}}{[}0,100)\\Tail \end{tabular}  & Acc@1 & Acc@5 \\
\midrule
baseline&0.499                              & 0.675                                                     & 0.553                                                        & 0.376                                                                    & 0.650                & 0.828 & 0.516                & 0.687                                                     & 0.568                                                        & 0.396                                                       & 0.663 & 0.837 \\
\midrule

LDAM + DRW&0.502                            & 0.680                                                     & 0.557                                                        & 0.378                                                                    & 0.656                & 0.811 & 0.518                & 0.687                                                     & 0.572                                                        & 0.397                                                       & 0.664 & 0.820 \\
EQL&0.502                                   & 0.679                                                     & 0.557                                                        & 0.378                                                                    & 0.653                & 0.829  & 0.518                & 0.690                                                     & 0.571                                                        & 0.398                                                       & 0.664 & 0.838 \\
CBS&0.491                                   & 0.649                                                     & 0.545                                                        & 0.371                                                                    & 0.640                & 0.820  & 0.507                & 0.660                                                     & 0.559                                                        & 0.390                                                       & 0.652 & 0.828 \\
CB Loss&0.495                                  & 0.653                                                     & 0.546                                                        & 0.381                                                                    & 0.643                & 0.823 &  0.511                & 0.665                                                     & 0.561                                                        & 0.398                                                       &   0.656    &  0.832 \\
mixup&0.484 & 0.649 & 0.535  & 0.368 & 0.633 & 0.818 & 0.495 & 0.660 &0.546	&0.381	&0.641	&0.824 \\\midrule

Ours&\textbf{0.516}&\textbf{0.683}&\textbf{0.569}&\textbf{0.397}&\textbf{0.658}&\textbf{0.834}& \textbf{0.532} &\textbf{0.695}& \textbf{0.584} &\textbf{0.417}&\textbf{0.667}& \textbf{0.843} \\

\bottomrule
\end{tabular}
\vspace{1mm}
\caption{Results and comparisons of different approaches for long-tailed recognition using features extracted from ResNet-50 and ResNet-101 (aggregated with a non-linear model), our method \system outperforms other long-tailed methods designed for image classification by clear margins.}
\vspace{-5mm}
\label{tab:2d-nonlinear}
\end{center}
\end{table*}

\section{Experiments}

\subsection{Settings}

\noindent
\textbf{Implementation Details.} During training,  we set the initial learning rate of the Adam optimizer to 0.0001 and decrease it every 30 epochs; we train for a maximum of 100 epochs by randomly sampling 60 frames as inputs, and the batch size is set to 128. At test time,  150 frames are uniformly sampled from raw features. For \system, we use a mix ratio $\eta$ to control how many samples are mixed in a mini-batch, and set $\eta$ to 0.5. In addition, the clip length of \system $L$ is set to 60.

\vspace{0.05in}
\noindent
\textbf{Backbone networks.}
To validate the generalization of our method for long-tailed video recognition, we follow the experimental settings as Decoupling~\cite{decoupling}. We use two popular backbones to extract features including: ResNet-101~\cite{resnet} pretrained on ImageNet and ResNet-50~\cite{resnet} pretrained on ImageNet. We also experiment with TSM~\cite{lin2019tsm} using a ResNet-50~\cite{lin2019tsm} as its backbone and the model is pretrained on Kinetics-400. We take the features from the penultimate layer of the network, resulting in features with a dimension of 2048.

We decode all videos at 1 fps, and then resize frames to the size of 256 and crop the center of them as inputs; these frames are uniformly sampled to construct a sequence with a length of 150. 

Note that we do not finetune the networks on VideoLT due to computational limitations. Additionally, we hope \system can serve as a plugin module to existing backbones with minimal surgery. Further, given features from videos, we mainly use a non-linear classifier with two fully-connected layers to aggregate them temporally. To demonstrate the effectiveness of features, we also experiment with NetVLAD for feature encoding~\cite{netvlad} with 64 clusters and the hidden size is set to 1024. 

\vspace{0.05in}
\noindent
\textbf{Evaluation Metrics.}
To better understand the performance of different methods for long-tailed distribution, we calculate mean average precision for head, medium and tail classes in addition to dataset-wise mAP, Acc@1 and Acc@5. Long-tailed video recognition requires algorithms to obtain good performance on tail classes without sacrificing overall performance, which is a new challenge for existing video recognition models.

\subsection{Results}

\noindent
\textbf{Comparisons with SOTA methods}.
We compare \system with three kinds of long-tailed methods that are widely used for image recognition tasks (See details and other extensions including square-root sampling and two-stage methods in the Supplementary material.):
\begin{itemize}
\item[$\bullet$] \textbf{Re-sampling}: We implement class-balanced sampling (CBS)~\cite{decoupling, ca-sampling} which uses equalized sampling strategies for data from different classes. In a mini-batch, it takes a random class and then randomly samples a video, and thus videos from head and tail classes share the same probability to be chosen.

\noindent
\item[$\bullet$] \textbf{Re-weighting}: It takes sampling frequency of each class into consideration to calculate weights for cross-entropy or binary cross-entropy. We conduct experiments with, Class-balanced Loss~\cite{cbloss}, LDAM Loss~\cite{ldam} and EQL~\cite{eql}.

\noindent
\item[$\bullet$] \textbf{Data augmentation}: We use the popular method Mixup~\cite{zhang2017mixup} for comparisons. For fair comparisons, mixup is performed in the feature space as \system. In particular, mixup mixes features from two videos in a mini-batch by summing their features frame-by-frame.

\end{itemize}

Table~\ref{tab:2d-nonlinear} summarizes the results and comparisons of different approaches on VideoLT.
We observe that the performance of tail classes are significantly worse compared to that of head classes using both ResNet-50 and ResNet-101 for all methods. This highlights the challenge of long-tailed video recognition algorithms. In addition, we can see popular long tail recognition algorithms for image classification tasks are not suitable for video recognition. Class-balanced sampling and class-balanced losses result in slightly performance drop, compared to the baseline model without using any re-weighting and re-sampling strategies; LDAM+DRW and EQL achieve comparable performance for overall classes and tail classes. For mixup, its performance is even worse compared to the baseline model, possibly due to the mixing between features makes training difficult. Instead, our approach achieves better results compared with these methods. In particular, when using ResNet-101 features, \system achieves an overall mAP of 53.2\%, which is 1.6\% and 1.4\% better compared to the baseline and the best performing image-based method (\ie, LDAM+DRW and EQL). Furthermore, we can observe that although CB Loss brings slightly better performance on tail classes, this comes at the cost of performance drop for overall classes. Compared to the CB Loss, \system significantly improves the performance by 2.1\% for tail classes without sacrificing the overall results.

\begin{table}[]
\renewcommand\arraystretch{1.1} 
\begin{center}
\renewcommand\tabcolsep{2.5pt}
\begin{tabular}{c|c|cccc}

\toprule
 & \multicolumn{1}{c|}{ LT Methods}   &  Overall &  \begin{tabular}[c]{@{}c@{}}{[}500,$+\infty$)\\ Head\end{tabular} & \begin{tabular}[c]{@{}c@{}} {[}100,500)\\ Medium\end{tabular} & \begin{tabular}[c]{@{}c@{}}{[}0,100)\\ Tail\end{tabular}              \\\midrule\midrule

\parbox[t]{5mm}{ \multirow{7}{*}{\rotatebox[origin=c]{90}{Nonlinear Model}} } & baseline                            & 0.565   & 0.757          & 0.620             & 0.436          \\
 & LDAM + DRW                               & 0.565   & 0.750          & 0.620             & 0.439     \\
 & EQL                                 & 0.567   & 0.757          & 0.623             & 0.439          \\
 & CBS                                 & 0.558   & 0.733          & 0.612             & 0.435          \\
 & CB Loss                             & 0.563   & 0.744          & 0.616             & 0.440          \\
 & Mixup                               & 0.548   & 0.736          & 0.602             & 0.425          \\
 & Ours                       & \textbf{0.580}  & \textbf{0.759}      & \textbf{0.632}      & \textbf{0.459}   \\
 \midrule\midrule

\parbox[t]{5mm}{ \multirow{7}{*}{\rotatebox[origin=c]{90}{NetVLAD Model}} } & baseline                            & 0.660   & 0.803          & 0.708             & 0.554        \\
& LDAM + DRW                               & 0.627   & 0.779          & 0.675      & 0.519   \\
& EQL                                 & 0.665   & \textbf{0.808}          & \textbf{0.713}            & 0.557    \\
& CBS                                  & 0.662   & 0.806          & 0.708             & 0.558        \\
& CB Loss                            & 0.666   & 0.801          & 0.712             & \textbf{0.566}      \\
& Mixup                               & 0.659   & 0.800          & 0.706             & 0.556          \\
& Ours                       & \textbf{0.667}   & 0.806          &\textbf{0.713}            & \textbf{0.566}  \\
\bottomrule
\end{tabular}
\vspace{1mm}
\caption{Results and comparisons using TSM (ResNet-50). Top: features aggregated using a non-linear model; Bottom: features aggregated using NetVLAD.}
\vspace{-5mm}
\label{tab:tsm}
\end{center}
\end{table}

\vspace{0.05in}
\noindent
\textbf{Extensions with more powerful backbones.} We also experiment with a TSM model using a ResNet-50 as its backbone to demonstrate the compatibility of our approach with more powerful networks designed for video recognition. In addition, we use two feature aggregation methods to derive a unified representations for classification. The results are summarized in Table~\ref{tab:tsm}. We observe similar trends as Table~\ref{tab:2d-nonlinear} using a nonlinear model---\system outperforms image-based long-tailed algorithms by 1.5\% and 2.3\% for overall and tail classes, respectively. In addition, we can see that features from the TSM model pretrained on Kinetics are better than image-pretrained features (58.0\% \vs 53.2\%). Furthermore, we can see that our approach is also compatible with more advanced feature aggregation strategies like NetVLAD. More specifically, with NetVLAD, \system outperforms the baseline approach by 0.7\% and 1.2\% for overall classes and tail classes, respectively.

\subsection{Discussion}

We now conduct a set of studies to justify the contributions of different components in our framework and provide corresponding discussion.

\vspace{0.05in}
\noindent
\textbf{Effectiveness of AP.} Throughout the experiments, we mainly use AP as a metric to adjust the number of frames used in \system. To test the effectiveness of AP, we also experiment with a constant $\beta$ and class frequency. The results are summarized in Table~\ref{tab:ap}. For a constant $\beta$, we test a variant of $\beta$ with $\beta=0.5$ throughout the training which takes the same number of frames for all classes. Results show the overall mAP on Nonlinear and NetVLAD model decrease at 10.2\% and 2.0\% respectively compared to the baseline, which suggests that sampling more frames for tail classes and fewer frames for head classes is a more practical strategy in long-tailed video recognition scenario.  Class frequency is another popular metric widely used for image-based long-tailed recognition~\cite{ca-sampling,mahajan2018exploring, bbn, decoupling}. In particular, we take the inverse frequency of each class to compute the number of frames sampled for head and tail classes, and then concatenate two clips as \system.  We observe that resampling videos with class frequency results in 0.5\% performance drop on the NetVLAD model. In contrast, using running average precision is a better way for resampling frames since it is dynamically derived based on knowledge learned by networks so far. As a result, it changes sampling rates based on the performance of particular classes, which prevents overfitting for top-performing classes and avoids under-fitting for under-performing categories at the same time. As aforementioned, treating weakly labeled videos as images and then resampling/reweighting them using class frequency might be problematic because some snippets might consist of  background frames. 

\begin{table}[]
\renewcommand\arraystretch{1.0} 
\begin{center}
\renewcommand\tabcolsep{2.5pt}
\begin{tabular}{c|c|cccc}
\toprule
Model    & \multicolumn{1}{c|}{\begin{tabular}[c]{@{}c@{}}CB\\ Strategy\end{tabular}} & \multicolumn{1}{c}{Overall} & \multicolumn{1}{c}{\begin{tabular}[c]{@{}c@{}}{[}500,$+\infty$)\\ Head\end{tabular}} & \multicolumn{1}{c}{\begin{tabular}[c]{@{}c@{}}{[}100,500)\\ Medium\end{tabular}} & \multicolumn{1}{c}{\begin{tabular}[c]{@{}c@{}}{[}0,100)\\ Tail\end{tabular}} \\\midrule\midrule
  & -   & 0.516   &  0.687 &  0.568  & 0.396 \\
  &  $\beta=0.5$  & 0.414   & 0.589   & 0.460           & 0.308  \\
Nonlinear & w/ CF       &   0.520 & 	0.680	& 0.571	& 0.405  \\
         & w/ rAP       & \textbf{0.532} &\textbf{0.695}& \textbf{0.584} &\textbf{0.417} \\\midrule
        &  -  & 0.668 & 0.775 & 	\textbf{0.707}	& 0.584 \\
        & $\beta=0.5$  & 0.648   & 0.758  & 0.684  & 0.567  \\
NetVLAD  & w/ CF       &  0.663	& 0.767	& 0.699	& 0.584	\\
         & w/ rAP       &\textbf{0.670}   & \textbf{0.780}                                                    & \textbf{0.707}                                                        & \textbf{0.590}                            \\\bottomrule                 
\end{tabular}
\vspace{1mm}
\caption{Results and comparisons of using running AP and other variants of $\beta$ to determine how many frames should be used from video clips.}
\label{tab:ap}
\end{center}
\end{table}

\begin{table*}[]
\renewcommand\arraystretch{1.2} 
\begin{center}

\renewcommand\tabcolsep{2.5pt}
\resizebox{1.0\linewidth}{!}{

\begin{tabular}{c|c|cccccc|cccccc}
\toprule
 & & \multicolumn{6}{c}{TSM (ResNet-50)}                                                                                                                                                                                  & \multicolumn{6}{|c}{ResNet-101}                                                                                                                                                                                \\\midrule
                        
                        & LT-Methods & Overall & \begin{tabular}[c]{@{}c@{}}{[}500,$+\infty$)\\ Head\end{tabular} & \begin{tabular}[c]{@{}c@{}}{[}100,500)\\ Medium\end{tabular} & \begin{tabular}[c]{@{}c@{}}{[}0,100)\\ Tail\end{tabular} & Acc@1 & Acc@5 & Overall & \begin{tabular}[c]{@{}c@{}}{[}500,$+\infty$)\\ Head\end{tabular} & \begin{tabular}[c]{@{}c@{}}{[}100,500)\\ Medium\end{tabular} & \begin{tabular}[c]{@{}c@{}}{[}0,100)\\ Tail\end{tabular} & Acc@1 & Acc@5 \\\midrule\midrule

\parbox[t]{5mm}{ \multirow{3}{*}{\rotatebox[origin=c]{90}{NonLinear}} } & baseline                & 0.565   & 0.757                                                     & 0.620                                                        & 0.436                                                    & 0.680 & 0.851 & 0.516   & 0.687                                                     & 0.568                                                        & 0.396                                                    & 0.663 & 0.837 \\
& FrameStack-BCE          & 0.568   & 0.751                                                     & 0.622                                                        & 0.445                                                    & 0.679 & 0.855 & 0.521   & 0.684                                                     & 0.571                                                        & 0.406                                                    & 0.660 & 0.839 \\
& FrameStack-FL           & \textbf{0.580}  & \textbf{0.759}                                                     & \textbf{0.632}                                                        &\textbf{0.459}                                                    & \textbf{0.686} &\textbf{0.859} & \textbf{0.532} &\textbf{0.695}& \textbf{0.584} &\textbf{0.417}&\textbf{0.667}& \textbf{0.843} \\\midrule\midrule

\parbox[t]{5mm}{ \multirow{3}{*}{\rotatebox[origin=c]{90}{NetVLAD}} } & baseline      & 0.660   & 0.803  & 0.708 & 0.554     & 0.695 & 0.870 & 0.668  & 0.775      & 0.707   & 0.584   & 0.700 & \textbf{0.864} \\
& FrameStack-BCE          & \textbf{0.669}   & \textbf{0.807}                                                    & \textbf{0.715}                                                        & \textbf{0.568}                                                    & \textbf{0.711} &\textbf{0.872} & \textbf{0.671}  & \textbf{0.781}       & 0.707   & 0.589       & 0.709 & 0.858 \\
& FrameStack-FL           & 0.667   & 0.806                                                     & 0.713                                                        & 0.566                                                    & 0.708 & 0.866 & 0.670  & 0.780                                                    & 0.707                                                       & \textbf{0.590}                                                 & \textbf{0.710} & 0.858 \\
\bottomrule
\end{tabular}}

\vspace{1mm}
\caption{Results of our approach using different loss functions and comparisons with baselines. \system is complemented with focal loss on Nonlinear model and TSM (ResNet-50), ResNet-101 features.}
\vspace{-5mm}

\label{tab:focal}
\end{center}
\end{table*}

\vspace{0.05in}
\noindent
\textbf{Effectiveness of loss functions.} As mentioned above, our approach resamples data from different classes and it is trained with focal loss. We now investigate the performance of our approach with different loss functions and the results are summarized in Table~\ref{tab:focal}. We observe that using \system is compatible with both loss functions, outperforming the baseline model without any resampling/reweighting strategies. For the nonlinear model, \system achieves better performance, while for NetVLAD, FrameStack with binary cross-entropy loss(BCE) is slightly better.

\begin{figure}[]
\begin{center}
\includegraphics[width=\linewidth]{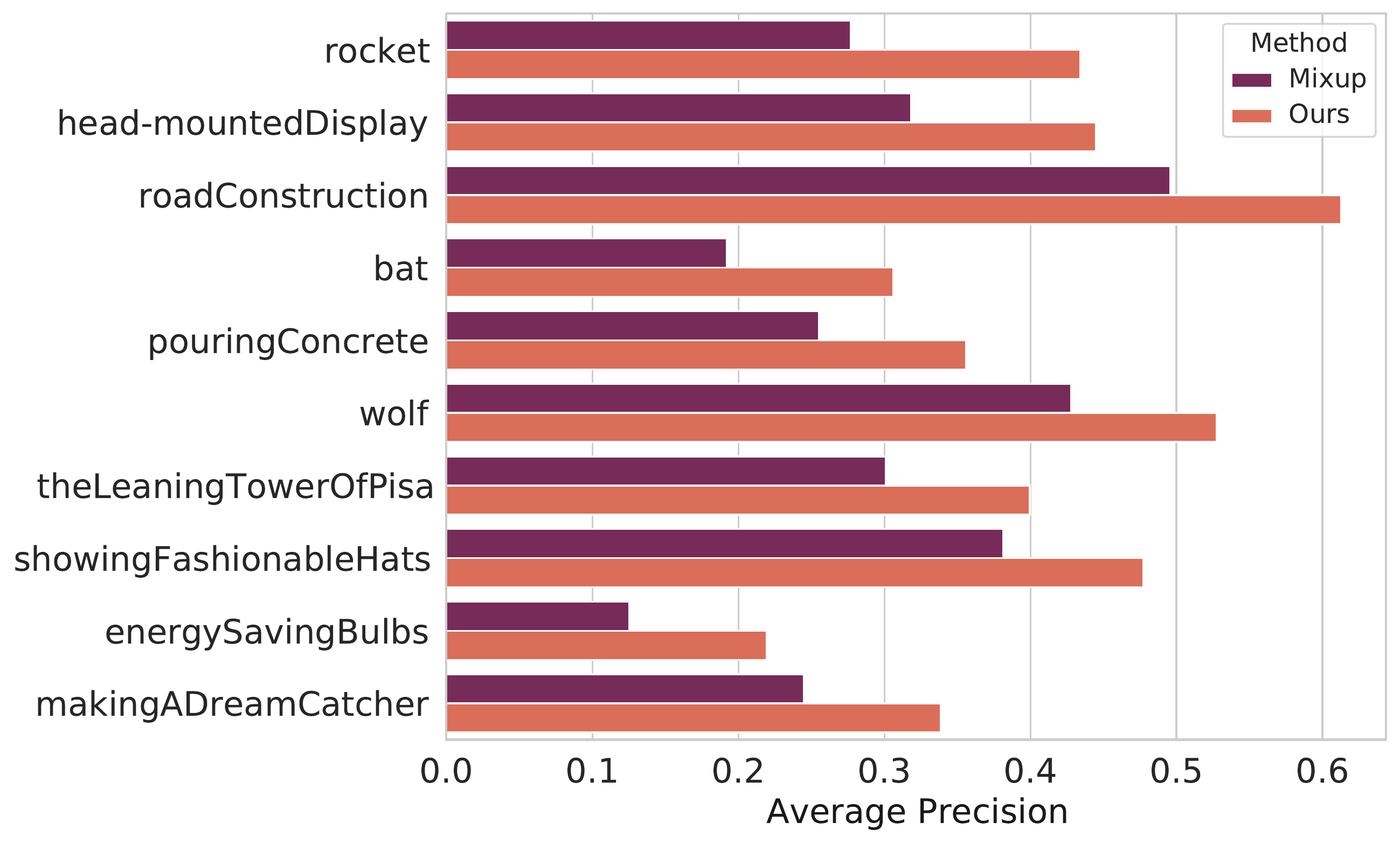}
\end{center}
  \caption{Top 10 among 1004 classes that \system surpasses Mixup. 40\% classes are action classes.}
\label{fig:compare-all}
\end{figure}

\begin{figure}[]
\begin{center}
\includegraphics[width=\linewidth]{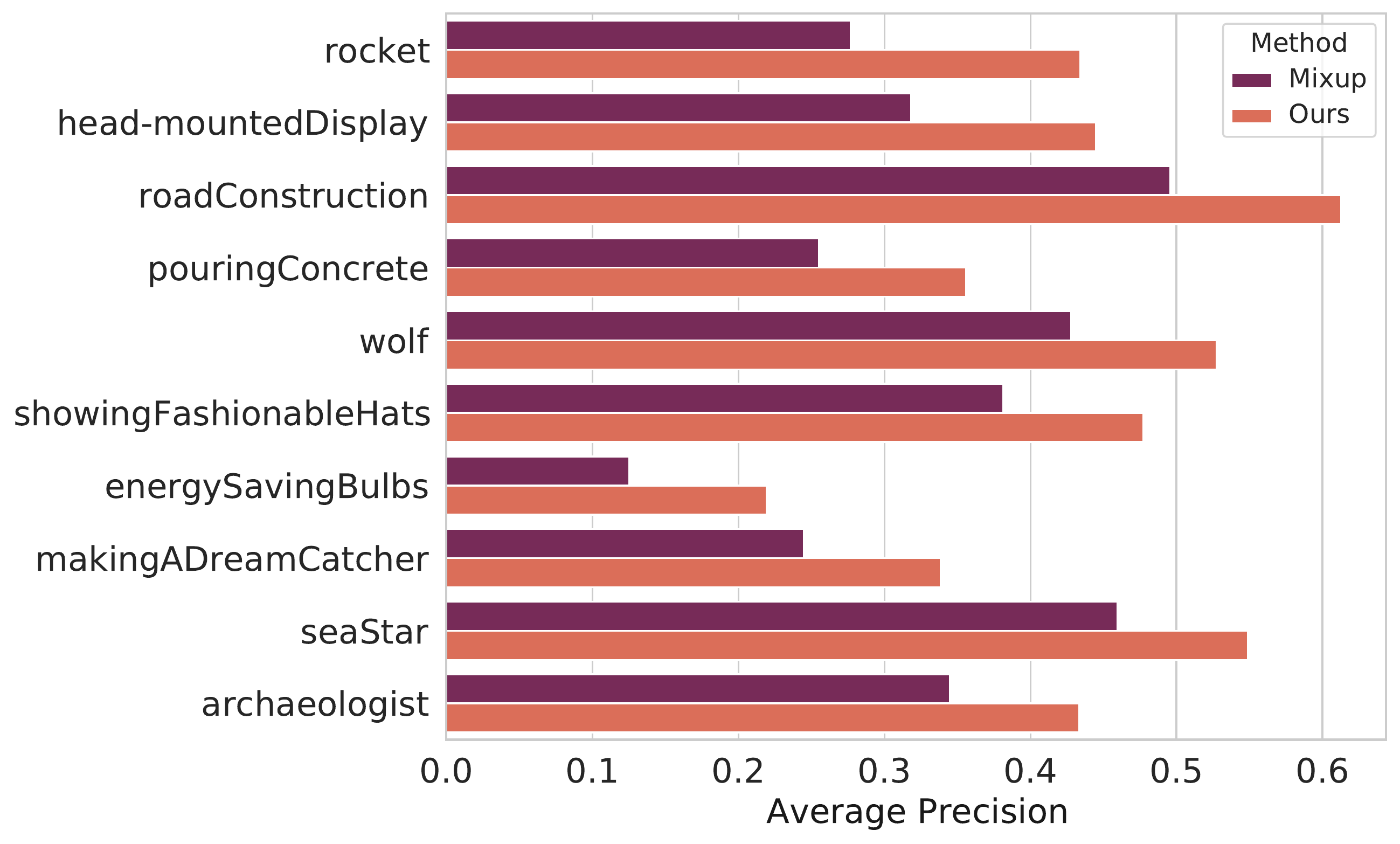}
\end{center}
  \caption{Top 10 among 340 tail classes that \system surpasses Mixup. Comparing across Figure~\ref{fig:compare-all}, we see \system achieves better performance mainly on tail clases.}
\label{fig:compare-tail}
\end{figure}

\vspace{0.05in}
\noindent
\textbf{Effectiveness of mixing ratio.}
We also investigate how many samples that are mixed in a mini-batch,  determined by a mixing ratio $\eta$, influence the performance. From Table~\ref{tab:ratio} we find that as $\eta$ increases, the performance of overall and tail classes increases at the beginning and then decreases---$\eta=0.5$ reaches highest performance which is adopted in our \system. It suggests that mixing all data in an epoch makes training more difficult.

\begin{table}[]
\renewcommand\arraystretch{1.0} 
\begin{center}

\renewcommand\tabcolsep{1pt}

\begin{tabular}{p{0.7cm}|cccccc}
\toprule
  $\eta$       & Overall              & \begin{tabular}[c]{@{}c@{}}{[}500,$+\infty$)\\ Head\end{tabular} & \multicolumn{1}{c}{\begin{tabular}[c]{@{}c@{}}{[}100,500)\\ Medium\end{tabular}} & \multicolumn{1}{c}{\begin{tabular}[c]{@{}c@{}}{[}0,100)\\ Tail\end{tabular}} & Acc@1                     & Acc@5                     \\\midrule\midrule

0        & 0.668 &	0.775	& 0.707 &	0.584  & 0.700	& \textbf{0.864}                  \\
0.3   &  0.667	& 0.780	& 0.707 &	0.586	& 0.710	  & 0.860   \\
0.5      & \textbf{0.670}	& \textbf{0.780}	& 0.707	&\textbf{0.590}& 0.710	& 0.858 \\
0.7      & 0.669	& 0.780	& 0.707	& 0.585	& 0.709	& 0.860 \\
0.9      & 0.668	& 0.774	& 0.704	& 0.588	& 0.706	& 0.859   \\\bottomrule
\end{tabular}
\vspace{1mm}
\caption{Effectiveness of the mixing ratio $\eta$, test results are based on ResNet-101 feature and NetVLAD Model.}
\vspace{-5mm}
\label{tab:ratio}
\end{center}
\end{table}

\vspace{0.05in}
\noindent
\textbf{FrameStack \vs Mixup.}
We compare the performance of FrameStack and Mixup for the overall and tail classes. Specifically, we compute the difference of average precision for each class between \system and Mixup. In Figure~\ref{fig:compare-all}, we visualize the top 10 classes from 1,004 classes that FrameStack outperforms Mixup and find that 40\% of them are action classes. When comparing Figure~\ref{fig:compare-all} and Figure~\ref{fig:compare-tail} together, we observe that the 80\% of the top 10 classes are tail classes, which shows FrameStack is more effective than Mixup especially in recognizing tail classes.

\vspace{-2mm}

\section{Conclusion}
This paper introduced a large-scale long-tailed video dataset---VideoLT with an aim to advance research in long-tailed video recognition. Long-tailed video recognition is a challenging task because videos are usually weakly labeled. Experimental results show existing long-tailed methods that achieve impressive performance in image tasks are not suitable for videos. In our work, we presented \system, which performs sampling at the frame level by using running AP as a dynamic measurement. \system adaptively selects different number of frames from different classes. Extensive experiments on different backbones and aggregation models show \system outperforms all competitors and brings clear performance gains on both overall and tail classes. Future directions include leveraging  weakly-supervised learning~\cite{wang2017untrimmednets, nguyen2018weakly}, self-supervised learning~\cite{han2020self, Xu_2019_CVPR} methods to solve long-tailed video recognition.

\vspace{-2mm}
\section*{Acknowledgement}

This work was supported in part by National Natural Science Foundation of China (\#62032006).

{\small
\bibliographystyle{ieee_fullname}
\bibliography{output}
}

\end{document}